%% file: main.tex
\documentclass[runningheads]{llncs}

\usepackage{eccv}

\usepackage{eccvabbrv}

\usepackage{graphicx}
\usepackage{booktabs}

\usepackage[accsupp]{axessibility}  % Improves PDF readability for those with disabilities.

\usepackage{hyperref}

\usepackage{orcidlink}

\definecolor{mblue}{rgb}{0.40, 0.67, 0.88}
\definecolor{mgreen}{rgb}{0.55, 0.78, 0.25}
\definecolor{mred}{rgb}{1., 0.39, 0.31}
\definecolor{mpurple}{rgb}{0.74, 0.50, 0.74}
\definecolor{morange}{rgb}{1.0, 0.55, 0.0}

\newcommand{\PartField}{PartField}

\begin{document}

\title{Interact3D: Compositional 3D Generation of Interactive Objects} 


\author{Hui Shan\inst{1,2,3}\orcidlink{0009-0000-8225-233X} \and % China
Keyang Luo\textsuperscript{\dag}\orcidlink{0009-0003-4491-8235} \and % no affiliation, China
Ming Li\inst{1,2,3}\orcidlink{0009-0000-5928-7542} \and % China
Sizhe Zheng\inst{2,3}\orcidlink{0009-0002-1457-8375} \and % China
Yanwei Fu\inst{2,5}\orcidlink{0000-0002-6595-6893} \and % China
Zhen Chen\inst{4}\orcidlink{0000-0002-6766-9046} \and % US
Xiangru Huang\inst{3}\thanks{Corresponding author.}\orcidlink{0000-0002-9533-9546}} % China

\authorrunning{H.~Shan et al.}

\institute{Zhejiang University, China \and
Shanghai Innovation Institute, China \and
Westlake University, China \and
Adobe, United States \and
Fudan University, China \\
\email{huangxiangru@westlake.edu.cn}
}

\maketitle
\begingroup
\renewcommand{\thefootnote}{\dag}
\footnotetext{Independent researcher from China.}
\endgroup

\begin{figure}[h] 
    \centering
    \includegraphics[width=0.83\linewidth]{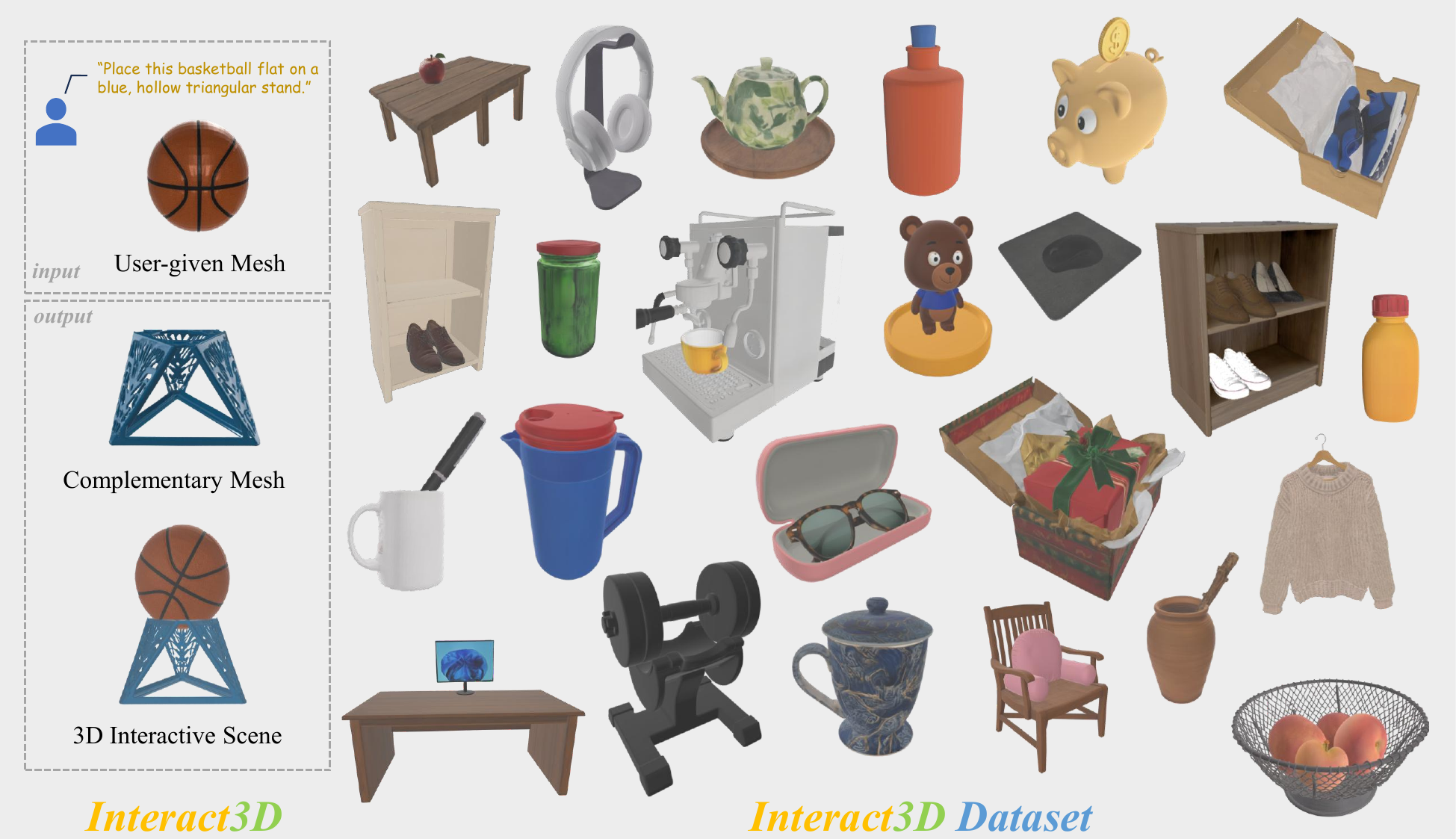}
    \caption{Given a text prompt and a user-given mesh, Interact3D synthesizes a high-quality, geometrically-compatible complementary mesh. It then seamlessly composes these two assets into an interactive 3D scene. Unlike existing approaches, our framework automatically generates collision-aware and physically-sound 3D environments.}
    \label{fig:teaser}
\end{figure}

\input{Section/0.abstract}
\input{Section/1.Intro}

\input{Section/2.RelatedWork}
\input{Section/3.Preliminary}
\input{Section/4.Methods}
\input{Section/5.Experiments}

\input{Section/6.Conclusions}
\bibliographystyle{splncs04}
\bibliography{main}

\appendix
\input{Section/Appendix}

\end{document}

%% file: Section/0.abstract.tex
\begin{abstract}
  Recent breakthroughs in 3D generation have enabled the synthesis of high-fidelity individual assets. However, generating 3D compositional objects from single images--particularly under occlusions--remains challenging. Existing methods often degrade geometric details in hidden regions and fail to preserve the underlying object-object spatial relationships (OOR). We present a novel framework \textbf{Interact3D} designed to generate physically plausible interacting 3D compositional objects. Our approach first leverages advanced generative priors to curate high-quality individual assets with a unified 3D guidance scene. To physically compose these assets, we then introduce a robust two-stage composition pipeline. Based on the 3D guidance scene, the primary object is anchored through precise global-to-local geometric alignment (registration), while subsequent geometries are integrated using a differentiable Signed Distance Field (SDF)-based optimization that explicitly penalizes geometry intersections. To reduce challenging collisions, we further deploy a closed-loop, agentic refinement strategy. A Vision-Language Model (VLM) autonomously analyzes multi-view renderings of the composed scene, formulates targeted corrective prompts, and guides an image editing module to iteratively self-correct the generation pipeline. Extensive experiments demonstrate that~\textbf{Interact3D} successfully produces promising collision-aware compositions with improved geometric fidelity and consistent spatial relationships. The code and dataset will be available at \url{https://github.com/SII-Hui/Interact3D}.
  \keywords{3D Generation \and Interaction \and Agentic Refinement}
\end{abstract}

%% file: Section/1.Intro.tex
\section{Introduction}
\label{sec:intro}
Robotic manipulation shows immense potential in domestic assistance, industrial automation, and disaster response. In recent years, learning-based manipulation paradigms, particularly Reinforcement Learning (RL)~\cite{ma2023eureka,yuan2025embodied,wu2023daydreamer}, have achieved remarkable progress. However, deploying these algorithms directly in the real world is expensive and unsafe. Therefore, training agents in physically simulated virtual interactive environments (Sim2Real) has become a dominant paradigm~\cite{makoviychuk2021isaac,narang2022factory,xu2023unidexgrasp}. To generalize to diverse real-world scenarios, these simulation platforms require a massive and diverse collection of interactive 3D assets with realistic geometry, plausible physical properties, and correct object–object relationships (OOR).

However, the availability of high-quality interactive 3D data remains severely limited. Curating these 3D datasets heavily relies on tedious manual modeling and annotation, making it difficult to scale. To bypass the scarcity of interactive 3D assets, recent works attempt to leverage large-scale human videos as an alternative supervision signal for robot learning~\cite{jain2024vid2robot, hoque2025egodex, zhang2026clap}. While video-driven approaches provide rich semantic priors and action trajectories, 2D videos inherently lack precise 3D geometric grounding and physical collision constraints. This fundamental limitation hinders the robot's ability to comprehend complex spatial relationships and perform fine-grained interactions in 3D space.

\begin{figure*}[t]
    \centering
    \includegraphics[width=\linewidth]{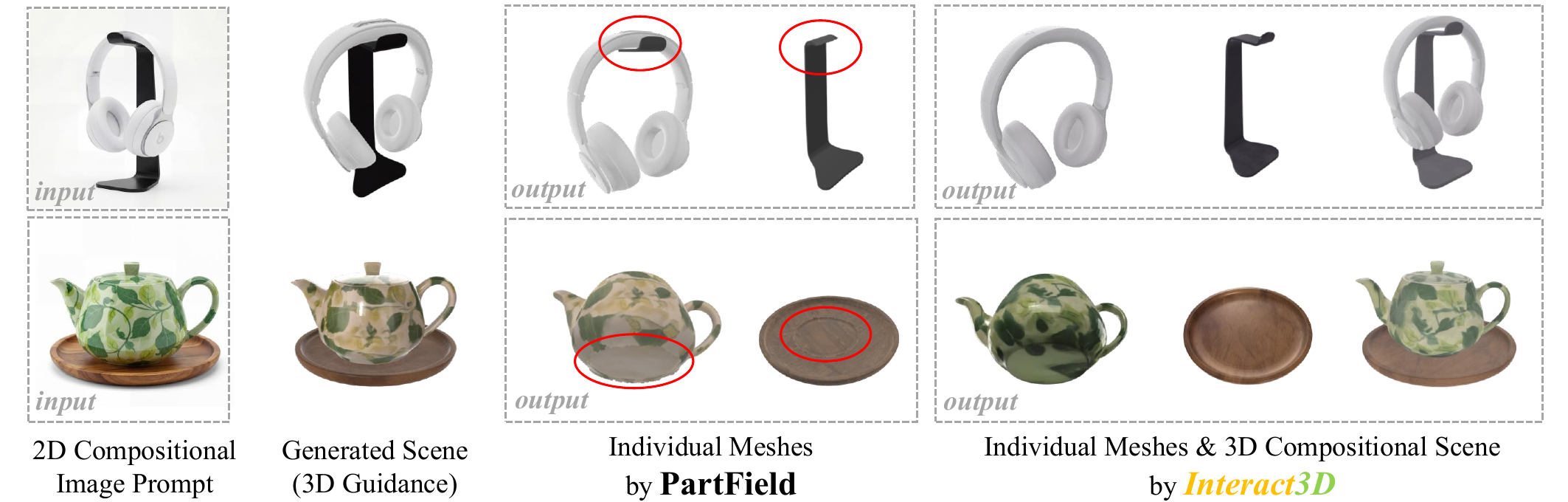}
    \caption{Given a generated scene, \textbf{\PartField} directly segments it to extract individual 3D assets, whereas \textbf{Interact3D} leverages it as 3D spatial guidance to compose high-quality independent geometries. (Though generating scenes and meshes via separate TRELLIS2 inferences yields minor texture variations, their spatial guidance remains robust to these discrepancies.)}
    \label{fig:Intro}
\end{figure*}

Concurrently, 3D generation models have achieved impressive results in synthesizing high-fidelity 3D assets from text or image prompts~\cite{xiang2025structured, zhao2025hunyuan3d, xiang2025native}. However, these models typically output monolithic and fused geometries which lack OOR. The generated meshes are ``baked'' into a single geometry, severely lacking independent geometry and physically meaningful spatial relationships. Naively segmenting such geometry, for instance, directly using PartField~\cite{liu2025partfield}, often introduces difficult-to-repair holes and inaccurate 3D assets (see Figure~\ref{fig:Intro}). Hunyuan3D~\cite{tencent_hunyuan3d_web} and Rodin~\cite{hyper3d_web} commercial websites can segment scenes and inpaint geometries, but suffer from the loss of the original texture in their remeshed geometry. Attempting to manually segment and optimize these monolithic assets is extremely labor-intensive and fails to generalize across diverse topologies. Therefore, the key challenge lies in how to reliably generate geometrically-compatible 3D compositional assets guided by user-given text or image prompts. An intuitive solution is to train a feed-forward network using large amounts of 3D compositional data~\cite{huang2025midi,yao2025cast}. 
However, feed-forward methods are inherently bottlenecked by the scarcity of 3D data, resulting in poor generalization and high training costs.

To address these challenges, we propose \textbf{Interact3D}, a novel pipeline for scalable and automated synthesis of interactive 3D scenes, see Figure~\ref{fig:overview}. Our key insight is to leverage the spatial priors implicitly encoded in image-guided 3D generation as guidance for geometric composition. By fully exploiting the priors of advanced 2D and 3D generative models, we reformulate compositional generation—traditionally requiring complex geometric reasoning—as a structured 3D registration problem.

However, unlike traditional 3D registration, compositional generation requires not only accurate alignment but also physically plausible interactions between objects. To this end, we first anchor the primary object using a global-to-local registration strategy, and then integrate additional components through a Signed Distance Field (SDF)-based optimization that explicitly penalizes interpenetrations, resulting in collision-aware compositions.

Moreover, inherent 2D occlusions often cause generated assets with complex joints to suffer from severe geometric mismatches, which cannot be fixed by simple rigid transformations. To address this, we propose a VLM-based agentic refinement strategy that formulates corrective text prompts to guide 2D generative image editing, thereby iteratively fixing the geometric flaws. Extensive experiments demonstrate the effectiveness of our methods in multi-object composition (see Figure~\ref{fig:teaser}). Our contributions are summarized in the following:
\begin{enumerate}
    \item We propose \textbf{Interact3D}, a generate-then-compose framework for training-free compositional 3D generation, featuring a two-stage, extensible composition pipeline, that enables the creation of collision-aware interactive 3D scenes.
    \item We introduce an interactive 3D dataset, which contains more than 8,000 interactive pairs and will be publicly released to facilitate future research.
    \item We develop a VLM-based agentic refinement strategy that automatically resolves severe geometric mismatches and improves compositional coherence.
\end{enumerate}

%% file: Section/2.RelatedWork.tex
\section{Related Work}
\label{sec:relatedwork}
\paragraph{\textbf{3D Generation.}} Recent advances in 3D generation have increasingly shifted away from slow, optimization-based paradigms~\cite{liu2024unidream,zhang2024dreammat,lin2023magic3d,chen2023fantasia3d,qiu2024richdreamer} to large-scale foundational models~\cite{zhang2024clay,zhao2025hunyuan3d,xiang2025structured,shan2025ni,xiang2025native,li2025step1x}, drastically improving both synthesis speed and geometric fidelity. CLAY~\cite{zhang2024clay} has advanced native 3D geometry generation, using 3D datasets from Objaverse~\cite{deitke2023objaverse} to produce highly controllable, realistic, and watertight assets. Hunyuan3D~\cite{zhao2025hunyuan3d} achieves industrial-grade asset quality by effectively unifying text-to-3D and image-to-3D pipelines within robust large-scale flow-based diffusion transformers. TRELLIS~\cite{xiang2025structured} introduces a unified structured latent representation (SLAT) and rectified flow transformers that significantly mitigate multi-view inconsistencies and artifacts in complex geometries. Building upon this foundation, its recent successor, TRELLIS2~\cite{xiang2025native} further scales the transformer architecture within the 3D latent space, achieving state-of-the-art in high-resolution 3D generation.
\paragraph{\textbf{3D Part Segmentation and Generation.}} The segmentation and generation of 3D parts, which are essential tasks in 3D generation, require a compositional understanding of the structures of 3D objects. SAMPart3D~\cite{yang2024sampart3d}, PartField, and P\textsuperscript{3}-SAM~\cite{ma2025p3} capture 3D features through feed-forward training and perform segmentation. Although they can segment parts generated by advanced 3D generation models, the resulting components often contain difficult-to-repair holes. In contrast, OmniPart~\cite{yang2025omnipart} and SAM3D~\cite{chen2025sam} are capable of directly generating complete 3D parts from masked images through expensive data curation and feed-forward training. Although the meshes they generate are watertight, their geometric accuracy, PBR materials, and alignment with image prompts still fall short when compared to 3D foundation models like TRELLIS2.
\paragraph{\textbf{3D Shape Composition.}} 3D shape composition~\cite{xiong2023learning,zakka2020form2fit} typically restores complete objects or scenes from multiple segmented components. Jigsaw~\cite{lu2023jigsaw}, NSM~\cite{chen2022neural} and RGL-NET~\cite{narayan2022rgl} focus on fragment composition without relying on predefined semantic information, as fragments often contain complex and precise geometric details. Based on these, Puzzlefusion~\cite{hossieni2023puzzlefusion}, Diffassemble~\cite{scarpellini2024diffassemble}, and Puzzlefusion++~\cite{wang2024puzzlefusion++} use diffusion models to refine the poses of fragments. 2BY2~\cite{qi2025two} is the first to propose the daily pairwise object composition task and contributes the corresponding dataset for the task. Compared to fragment composition, daily pairwise components always lack clear geometric relationships. Therefore, COPY-TRANSFORM-PASTE~\cite{gatenyo2026copy} utilizes text-guided optimization with vision-language models to supervise composition and achieves promising results. Although 2BY2 introduces a daily composition dataset, it consists of only 517 pairs, most of which are selected from existing 3D datasets~\cite{deitke2023objaverse,xiang2020sapien}. In contrast, we propose a fully automated pipeline for generating large-scale brand-new 3D composition datasets.

%% file: Section/3.Preliminary.tex
\section{Preliminary}
\label{sec:Preliminary}
Our method is designed for physically-sound 3D compositional generation. In this section, we first formally define the task of 3D compositional generation (section~\ref{problemsetup}), then briefly formulate 3D point cloud registration task and introduce one classical solution, Iterative Closest Point (ICP) (section~\ref{registration}).

\subsection{Problem Setup}
\label{problemsetup}
Given a 3D mesh $\mathbf{M} = (\mathbf{V}, \mathbf{F})$, where $\mathbf{V} \in \mathbb{R}^{N \times 3}$ denotes the vertex positions and $\mathbf{F} \in \mathbb{N}^{N_f \times 3}$ denotes the face indices, together with a text prompt that specifies compositional semantics, our goal is to generate a complementary 3D component $\mathbf{M}_{\mathrm{comp}}$ that forms a coherent compositional scene with $\mathbf{M}$.

In addition to synthesizing $\mathbf{M}_{\mathrm{comp}}$, we optimize the transformation parameters $\boldsymbol{\theta} = (\boldsymbol{\tau}, \mathbf{R}, \mathbf{s})$ between both meshes to achieve geometrically compatible composition, where $\boldsymbol{\tau} \in \mathbb{R}^3$ denotes translation, $\mathbf{R} \in \mathcal{SO}(3)$ denotes rotation, and $\mathbf{s} \in \mathbb{R}^+$ denotes uniform scaling.

\subsection{3D Point Clouds Registration}
\label{registration}

Given source and target point clouds 
$\mathcal{P} = \{\mathbf{p}_i\}_{i=1}^{N}$ and 
$\mathcal{Q} = \{\mathbf{q}_j\}_{j=1}^{M}$, 
we adopt \emph{scale-aware ICP}~\cite{ying2009scale} for similarity registration. The objective jointly estimates a uniform scale 
$s \in \mathbb{R}^+$, rotation $\mathbf{R} \in \mathcal{SO}(3)$, and translation $\boldsymbol{\tau} \in \mathbb{R}^3$:

\begin{equation}
\min_{\mathbf{s},\mathbf{R},\boldsymbol{\tau}}
\sum_{(\mathbf{p}_i,\mathbf{q}_j)\in C}
\left\| \mathbf{s}\cdot \mathbf{R}\cdot\mathbf{p}_i + \boldsymbol{\tau} - \mathbf{q}_j \right\|_2^2,
\label{eq:scale-icp}
\end{equation}

where $C$ denotes point correspondences. At each iteration, correspondences are updated using the scheme mentioned in~\cite{ying2009scale}, and the optimal similarity transformation $(\mathbf{s},\mathbf{R},\boldsymbol{\tau})$ is computed in closed form using Umeyama’s method~\cite{umeyama2002least}.

%% file: Section/4.Methods.tex
\section{Method}
\label{sec:Method}

We present a comprehensive framework for generating physically-sound 3D shape components. After curating high-quality individual assets and a 3D guidance scene, we process them through a novel two-stage composition pipeline (Section~\ref{DataCuration}). A global-to-local geometric alignment (Section~\ref{Global-to-local}) is used to accurately anchor the reference object in stage one. A SDF-based physics-aware optimization (Section~\ref{SDFBased}) is introduced to align subsequent objects while avoiding spatial intersections in stage two. For cases of severe and unavoidable collisions, we integrate an agentic, VLM-driven refinement loop (Section~\ref{Agentic}) to iteratively edit the scene until a low-collision configuration is achieved. The main notations are included in Table~\ref{tab:notations}.
\input{tables/notation_tab}

\subsection{Data Curation and Workflow}
\label{DataCuration}
We begin by rendering the input mesh $\mathbf{M}$ from a canonical frontal viewpoint, obtaining an image $I_\mathrm{rendered}$. Conditioning on this rendered image and a user-provided textual prompt, we use \emph{Nano Banana Pro}~\cite{gemini_web} to generate a compositional scene image $I_{\mathrm{scene}}$. To obtain the complementary component, we further prompt the model to remove the original object from $I_{\mathrm{scene}}$, producing an image $I_{\mathrm{comp}}$ that contains only the newly introduced geometry. Intuitively, $I_{\mathrm{comp}}$ captures the complementary part needed to complete the composition.

\begin{figure*}[t]
    \centering
    \includegraphics[width=0.95\linewidth]{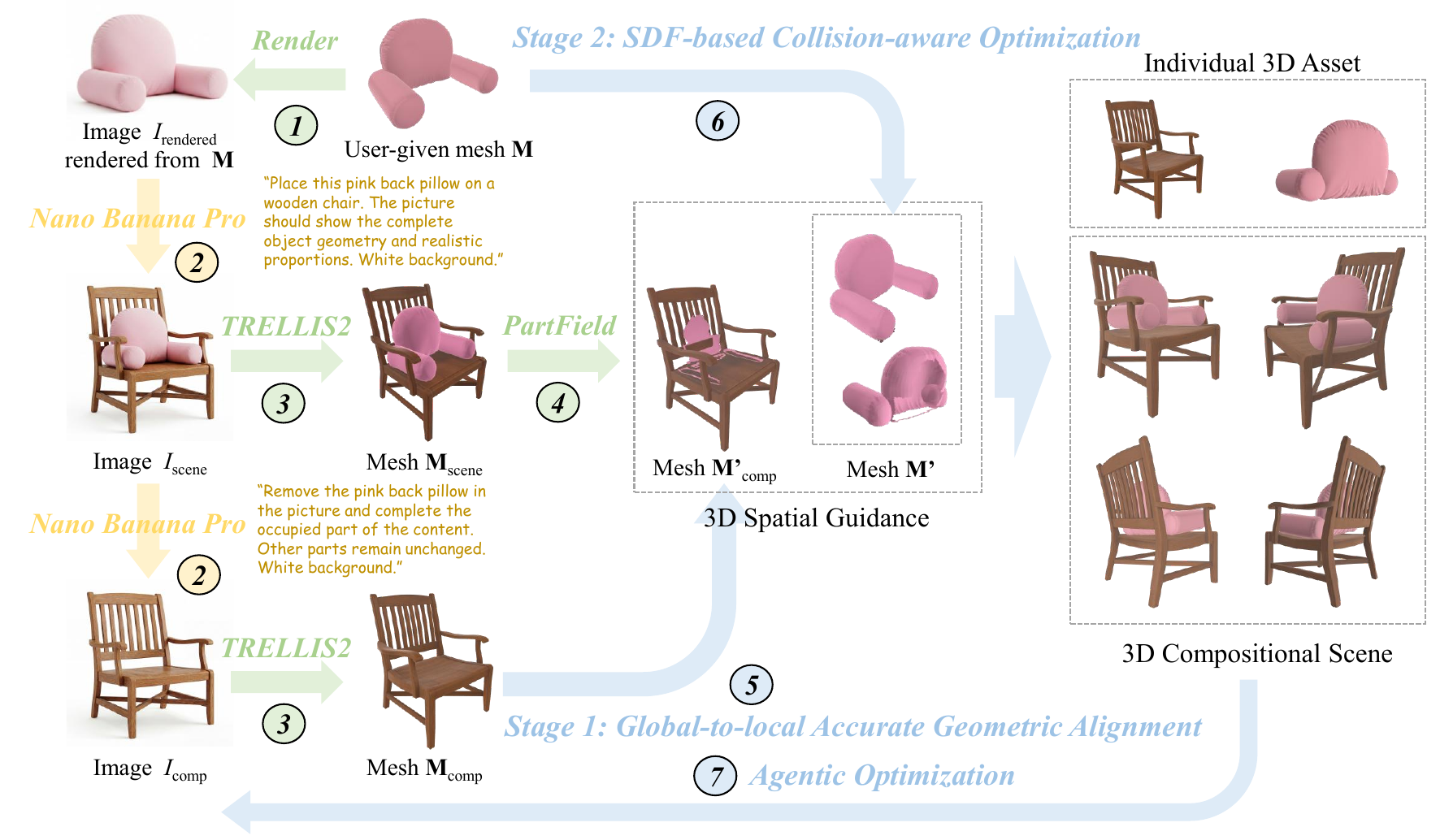}
    \caption{
    \textbf{Overview of Interact3D.} Given a user-provided mesh $\mathbf{M}$, we render it and use Nano Banana Pro to synthesize a guided scene image $I_\mathrm{scene}$ and a complementary image $I_\mathrm{comp}$. TRELLIS2 then reconstructs these into 3D meshes ($\mathbf{M_\mathrm{scene}}$ and $\mathbf{M_\mathrm{comp}}$) to provide spatial guidance. Relying on coarse parts segmented by PartField, we execute a two-stage composition. Stage 1 performs a global-to-local registration on the ``largest'' mesh to establish the anchor pose, while Stage 2 applies an SDF-based collision-aware optimization on the remaining mesh to resolve spatial intersections. (In this case, $\mathbf{M_\mathrm{comp}}$ is considered as $\mathbf{M_\mathrm{anchor}}$, $\mathbf{M}$ is considered as $\mathbf{M_\mathrm{remain}}$.) Finally, a VLM-based agentic refinement handles unavoidable collisions, yielding a physically-sound interactive 3D scene.
    }
    \label{fig:overview}
\end{figure*}

Subsequently, we use \emph{TRELLIS2} to generate $\mathbf{M_{\mathrm{scene}}}$ and $\mathbf{M_\mathrm{comp}}$ meshes separately based on images $I_\mathrm{scene}$ and $I_\mathrm{comp}$. To recover the spatial relationship between components, we apply \emph{\PartField{}} to segment $\mathbf{M_\mathrm{scene}}$ into two parts: $\mathbf{M'}$ and $\mathbf{M'_\mathrm{comp}}$. Although these segmented meshes often contain holes and geometric artifacts that make them unsuitable as final assets, they still provide reliable spatial cues. In particular, the transformation  $\boldsymbol{\theta}$ extracted from $\mathbf{M'}$ and $\mathbf{M'_{\mathrm{comp}}}$ serves as geometric guidance for composing the original mesh $\mathbf{M}$ and the high-quality complementary mesh $\mathbf{M_{\mathrm{comp}}}$ through 3D point cloud registration (see Figure~\ref{fig:overview}).

\subsubsection{Two-stage Composition Pipeline.} 
\label{subsec:two-stage-comp-pipelin}
We decompose composition into two sequential stages with distinct roles.
\paragraph{Stage 1: Anchor Alignment.} We first select an anchor object $\mathbf{M}_\mathrm{anchor}$ from the candidate meshes ($\mathbf{M}$ or $\mathbf{M}_{\mathrm{comp}}$). Empirically, the object with the larger projected area (measured by 2D bounding box size) in $I_{\mathrm{scene}}$ provides a more stable reference.  We apply the global-to-local registration strategy (Section~\ref{Global-to-local}) exclusively to this anchor 
object to recover an accurate initial pose with respect to the guidance mesh $\mathbf{M}'_{\mathrm{anchor}}$.
\paragraph{Stage 2: Collision-aware Composition.} Once the anchor object is fixed, we optimize the pose of the remaining component(s) $\mathbf{M}_\mathrm{remain}$ using the SDF-based collision-aware formulation (Section~\ref{SDFBased}) with the guidance $\mathbf{M}'_{\mathrm{remain}}$. This stage explicitly balances geometric alignment with collision avoidance, ensuring physically plausible composition.

\subsection{Global-to-local Accurate Geometric Alignment}
\label{Global-to-local}
This stage is applied only to the selected anchor object mentioned in Section~\ref{subsec:two-stage-comp-pipelin}. Its objective is to recover a reliable initial pose under partial overlap, occlusion, and reconstruction noise.

In practice, the generated compositional image $I_\mathrm{scene}$ inevitably introduces occlusions, and the individual meshes are not perfectly consistent with $\mathbf{M_\mathrm{scene}}$. Moreover, the subsequent segmentation process in \PartField{} often produces structural holes and incomplete surfaces (see Figure~\ref{fig:Intro}). As a result, the resulting point cloud pairs ($\mathbf{M}_\mathrm{anchor}$ and $\mathbf{M'_\mathrm{anchor}}$) typically exhibit low-overlap ratios and contain a significant number of outlier points. Under such conditions, traditional local registration methods, such as ICP, are highly sensitive to initialization and may converge to undesirable local minima.

To effectively address the aforementioned challenges, we adopt a robust and accurate global-to-local registration paradigm. We begin by resolving the scale discrepancies using an Oriented Bounding Box (OBB) approach to estimate the geometric extents and compute the initial scale factor $\mathbf{s}$. Following this scale alignment, we employ GeoTransformer~\cite{qin2023geotransformer}, a transformer-based 3D registration network, which is exceptionally robust in handling geometries with low-ratio overlaps. This step provides reliable global estimates of the initial global translation $\boldsymbol{\tau}$ and rotation $\mathbf{R}$. Finally, with these global estimates of $(\mathbf{s}, \mathbf{R}, \boldsymbol{\tau})$ as initialization, a scale-aware ICP algorithm is deployed to achieve precise alignment. In summary, this global initialization provides a warm start that prevents the algorithm from falling into local minima, allowing the subsequent local optimization to focus on minimizing residual geometric discrepancies for an accurate final alignment.

\subsection{SDF-Based Collision-Aware Composition}
\label{SDFBased}

After global-to-local registration, the primary object is fixed as the spatial anchor. The remaining components $\mathbf{M_\mathrm{remain}}$ are then placed relative to this anchor. While we can use a similar global-to-local alignment method for the rest of the components to provide a reasonable pose estimate, it does not guarantee physically valid placement. Small geometric discrepancies may lead to interpenetration or unnatural spacing between objects.

To refine the placement, we introduce a SDF-guided optimization stage. Given the anchor mesh $\mathbf{M}_\mathrm{anchor}$, we precompute its Signed Distance Field (SDF) $\Phi_{\mathrm{anchor}}(\mathbf{p})$, which encodes the signed distance from a query point $\mathbf{p}$ to the surface of $\mathbf{M}_\mathrm{anchor}$ and negative values indicate penetration into the anchor object.

Following our global-to-local approach, we first obtain the initial transformation $\boldsymbol{\theta}$ between $\mathbf{M}_\mathrm{remain}$ and $\mathbf{M}'_\mathrm{remain}$ using OBB  and GeoTransformer as a starting point. For any point $\mathbf{p}$ in $\mathbf{M}_{\mathrm{remain}}$, we define the collision loss $\mathcal{L}_\mathrm{col}$ using the ReLU operator $[\cdot]_{+}$ after transformation $\boldsymbol\theta$: 
\begin{equation}
\label{Equation:collision}
\mathcal{L}_\mathrm{col}(\boldsymbol\theta)
= \sum\limits_{\mathbf{p}\in\mathbf{M}_\mathrm{remain}} \left( \underbrace{\left[ -\Phi_{\mathrm{anchor}}(\boldsymbol\theta(\mathbf{p})) \right]_+^2}_{\text{Hard Penalty}} + \lambda \cdot \underbrace{\left[ \epsilon - \Phi_{\mathrm{anchor}}(\boldsymbol{\theta}(\mathbf{p})) \right]_+}_{\text{Soft Repulsion}} \right).
\end{equation}
Here, $\epsilon$ denotes the safety margin and $\lambda$ is set to a minimal value to ensure a smooth transition of the penalty field near the boundary of the object. The final placement is obtained by 
minimizing:
\begin{equation}
\label{Equation:Allloss}
\min_{\boldsymbol\theta} 
\; \sum\limits_{p\in\mathbf{M}_\mathrm{remain}}\|\boldsymbol{\theta}(\mathbf{p}) - \mathbf{p}'\|^2 
+ \beta^{(k)}\mathcal{L}_\mathrm{col}(\boldsymbol\theta),
\end{equation}
where the first term preserves the alignment between $\mathbf{M}_\mathrm{remain}$ and $\mathbf{M}_\mathrm{remain}'$, 
and $\beta^{(k)}$ balances geometric consistency and collision avoidance. Specifically, $\beta^{(k)}$ is initialized as 0, so that it is equivalent to the scale-aware ICP solver at the beginning and increases linearly to $\beta_{max}$ throughout the optimization process. At the $k$-th optimization stage, $\mathbf{p}'$ is first chosen from $\mathbf{M'_\mathrm{remain}}$ for each $\boldsymbol{\theta}(\mathbf{p})$ following~\cite{ying2009scale}, and then optimize Equation~\eqref{Equation:Allloss} with $\mathbf{p}'$ fixed. This schedule allows the optimization to prioritize global geometric alignment in the early stages and progressively enforce physical plausibility as the pose converges.

\begin{figure*}[t]
    \centering
    \includegraphics[width=\linewidth]{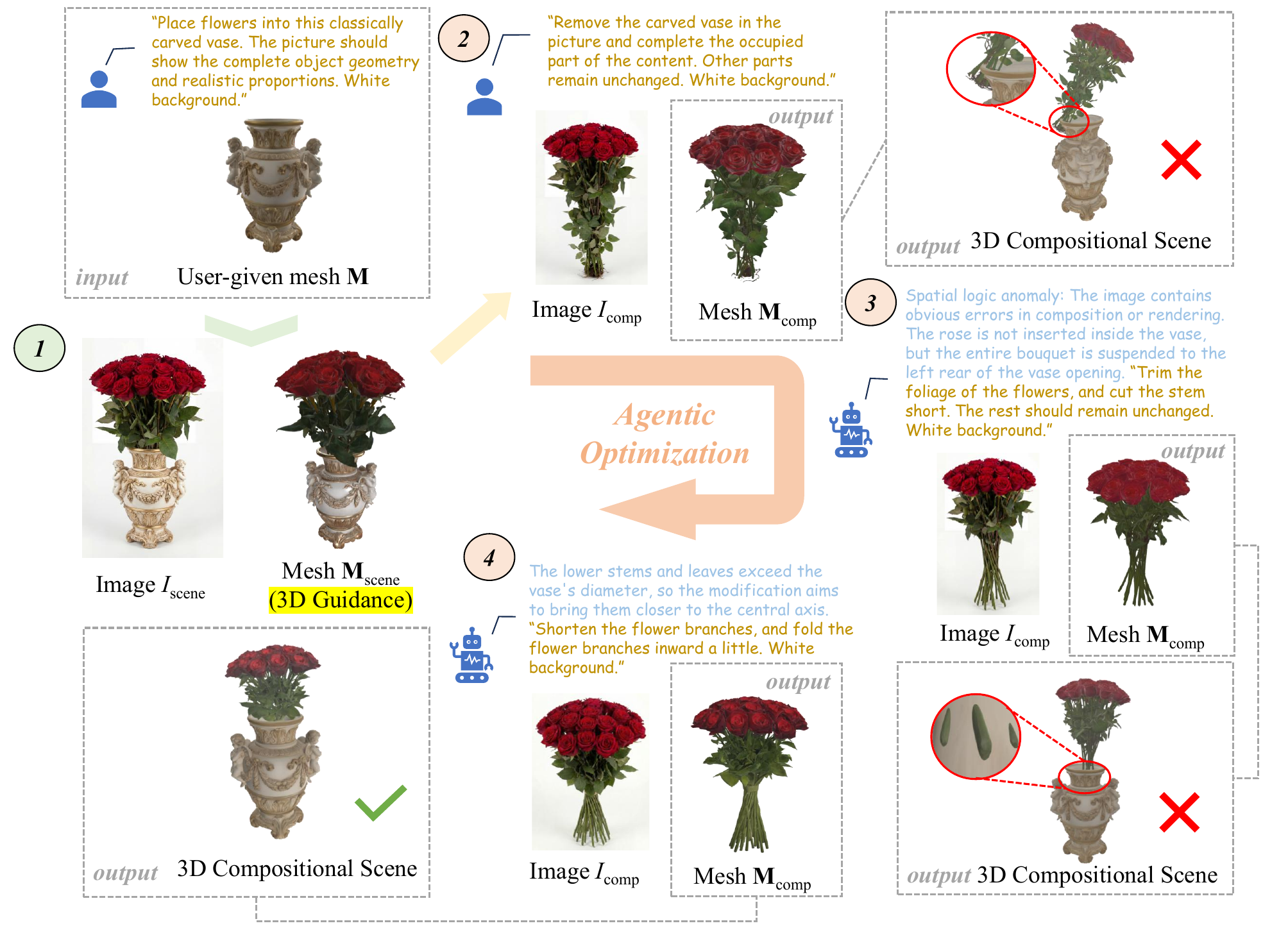}
    \caption{\textbf{Agentic Refinement.} When generating flowers inside a vase, severe occlusion in image $I_\mathrm{scene}$ causes the loss of object-object spatial relationships during TRELLIS2 reconstruction. As a result, the flower stem in the complementary image $I_\mathrm{comp}$ (generated by Nano Banana Pro) may not align correctly with the vase mesh $\textbf{M}$. This leads to geometric intersections after composition (see \emph{top-right} image). To resolve this, we render multi-view images, including internal cross-sections. These renderings are analyzed by a VLM, which generates a corrective text prompt to update the complementary image via Nano Banana Pro. This process continues until no more geometric intersections are found or the maximum iteration limit is reached.}
    \label{fig:AgenticOptimization}
\end{figure*}

By treating the anchor as a fixed spatial reference and refining the remaining components through SDF-guided optimization, our method generates geometrically compatible and physically plausible composition.

\subsection{Agentic Optimization}
\label{Agentic}

Although SDF-based optimization effectively resolves moderate  geometric discrepancies, it cannot correct fundamentally incompatible geometries (See Figure~\ref{fig:AgenticOptimization}). In practice, the complementary mesh generated from 2D guidance may deviate significantly from the intended spatial configuration, leading to persistent collisions or implausible object relationships that cannot be resolved through transformation alone.

To address this limitation, we introduce an agentic refinement 
mechanism that operates at the semantic level. Instead of further adjusting pose parameters, we revise the complementary geometry itself. Specifically, we render the current compositional scene from multiple viewpoints and feed the rendered images, along with the original compositional prompt, into a Vision-Language Model (VLM, e.g., Gemini 3 Pro~\cite{gemini_web}). The VLM analyzes spatial inconsistencies and generates corrective feedback in the form of targeted editing instructions. These instructions are then used to guide the 2D image editing model to refine the complementary component. The updated 2D image is subsequently reconstructed into a new 3D mesh and re-integrated into the composition pipeline.

This closed-loop process enables semantic-level correction beyond purely geometric optimization. By iteratively combining geometric alignment and language-guided refinement, our framework can resolve severe mismatches that are otherwise difficult to handle with traditional registration or SDF-based methods alone.

%% file: tables/notation_tab.tex
\begin{table}[ht]
\centering
\small
\caption{Summary of main notations.}
\label{tab:notations}
\begin{tabular}{l|l}
\toprule
\textbf{Symbol} & \textbf{Description} \\
\midrule

$\mathbf{M}, I_\mathrm{rendered}$
& Input 3D mesh, and corresponding rendered image \\

$I_{\mathrm{scene}}, I_{\mathrm{comp}}$
&Generated compositional and complementary image \\

$\mathbf{M}_{\mathrm{scene}}, \mathbf{M}_{\mathrm{comp}}$
& Scene and complementary mesh reconstructed from $I_\mathrm{scene}$ and $I_{\mathrm{comp}}$ \\

$\mathbf{M}_{\mathrm{anchor}}, \mathbf{M}_{\mathrm{remain}}$
& Anchor mesh and remaining mesh selected from 
$\{\mathbf{M}, \mathbf{M}_{\mathrm{comp}}\}$ \\

$\mathbf{M}', \mathbf{M}'_{\mathrm{comp}}$
& Segmented meshes extracted from $\mathbf{M}_{\mathrm{scene}}$ \\

$\mathbf{M}'_{\mathrm{anchor}}, \mathbf{M}'_{\mathrm{remain}}$
& Segmented counterparts from $\mathbf{M}_{\mathrm{scene}}$ \\

$\boldsymbol{\theta}(\mathbf{p})=\mathbf{s} \cdot \mathbf{R}\mathbf{p} + \boldsymbol{\tau}$
& Transformation function in terms of translation $\boldsymbol{\tau}$, rotation $\mathbf{R} $, scale $\mathbf{s}$ \\

$\Phi_{\mathbf{M}}(\mathbf{p})$
& Signed distance from point $\mathbf{p}$ to mesh $\mathbf{M}$ \\

\bottomrule
\end{tabular}
\end{table}

%% file: Section/5.Experiments.tex
\section{Experiments}
\label{sec:Exp}

\begin{figure*}[t]
    \centering
    \includegraphics[width=0.8\linewidth]{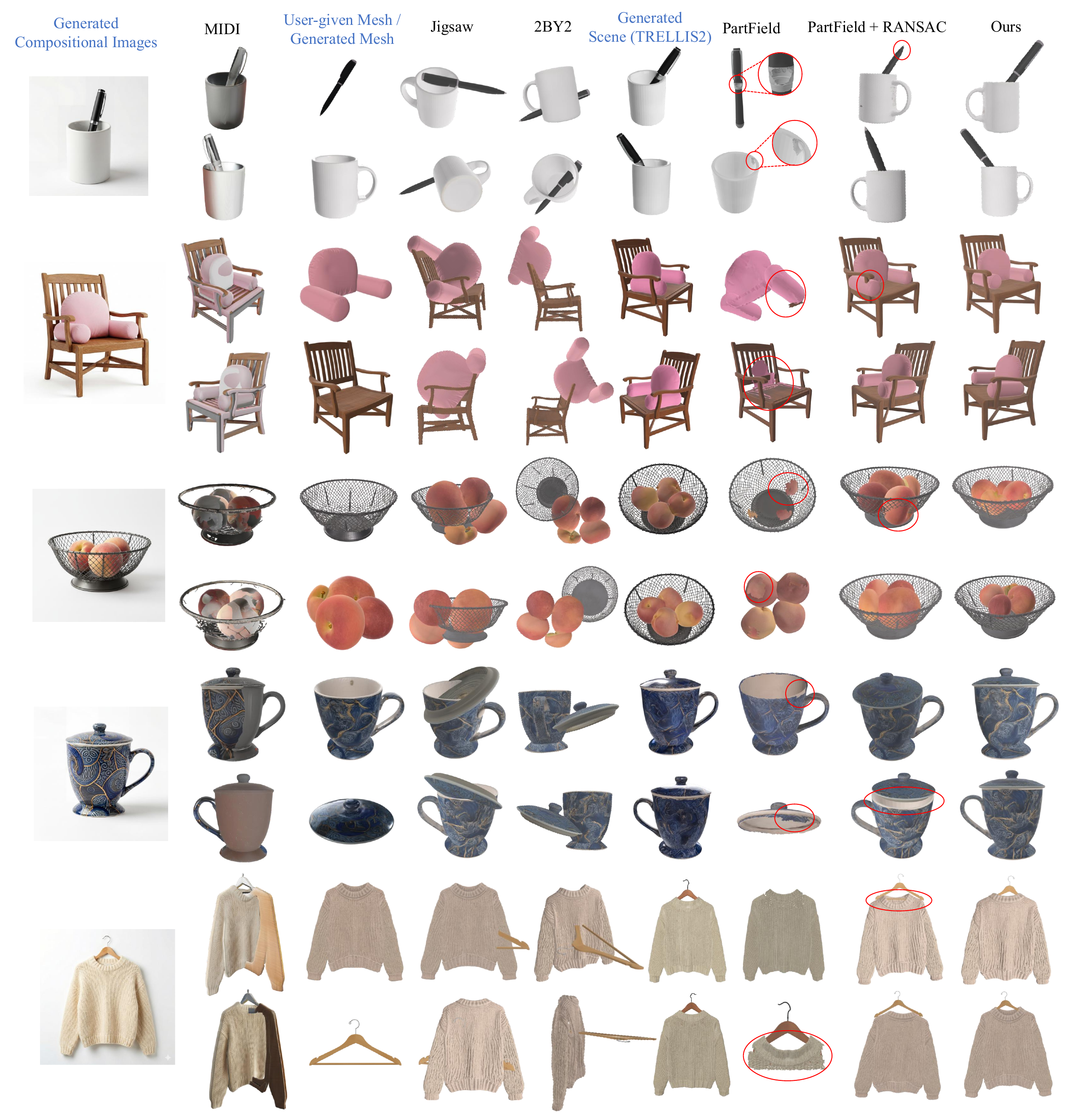}
    \caption{\textbf{Results of two-part composition.} Data-driven baselines (MIDI, Jigsaw, 2BY2) struggle to generalize to novel inputs. 3D segmentation of generated scenes (TRELLIS2 + PartField) inherently produces severe geometric holes (circled regions). Furthermore, ``PartField + RANSAC'' suffers from orientation inversion (1st and 2nd rows) and severe geometry intersections (circled region in the corresponding result column). In contrast, our framework consistently synthesizes physically plausible, and collision-aware components.}
    \label{fig:QualitativeResults}
\end{figure*}

\subsection{Experimental Setup}

We use Nano Banana Pro to generate images and adopt TRELLIS2 (and partially Hunyuan3D) for image-conditioned 3D generation. Our proposed framework is entirely training-free, and the inference process is conducted on a single NVIDIA H200 GPU. Specifically, we set $\lambda = 0.003$ to smooth the penalty field in Equation~\eqref{Equation:collision}, set $k_{max} = 100, \beta_{max} = 3.0$ in Equation~\eqref{Equation:Allloss} to progressively enforce physical plausibility. Furthermore, we set the maximum number of iterations in agentic optimization to 5. We empirically find our method to be robust to these hyperparameters within a reasonable range.

\paragraph{\textbf{Baselines.}} We evaluate the compositional performance of Interact3D against three categories of baselines.

\begin{enumerate}
    \item \textbf{Feed-forward baselines.} We compare our approach with Jigsaw~\cite{lu2023jigsaw} (for fractured object assembly), 2BY2~\cite{qi2025two} (for daily pairwise object composition), and MIDI~\cite{huang2025midi}. Specifically, Jigsaw and 2BY2 take the user-provided mesh $\mathbf{M}$ and the TRELLIS2-generated complementary mesh $\mathbf{M}_{\mathrm{comp}}$ as input, without additional image or text conditioning. Notably, both are inherently constrained by the inability to optimize the object scale $\boldsymbol{s}$. We adopt the official pretrained weights for Jigsaw and train 2BY2 from scratch following its official protocol. In contrast, MIDI serves as an image-conditioned baseline, which predicts the constituent 3D assets and their spatial positions from a compositional scene image.
    \item \textbf{Direct segmentation baseline.} To evaluate the impact of our generate-then-compose design, we construct a baseline that directly segments the TRELLIS2-generated scene mesh $\mathbf{M}_{\mathrm{scene}}$ using PartField. This approach does not require an extra pose estimation and is based solely on object-level segmentation. However, the resulting meshes typically contain geometric holes that are difficult to repair, inevitably altering the user-provided assets. Comparison against this baseline highlights the advantage of our method in preserving high-fidelity individual geometries.
    \item \textbf{Classical registration baseline.} To further demonstrate the effectiveness of our proposed registration approach (Section~\ref{Global-to-local} and \ref{SDFBased}), we employ the RANSAC~\cite{fischler1981random} algorithm to independently register the two inputs, $\mathbf{M}$ and $\mathbf{M_\mathrm{comp}}$, based on $\mathbf{M'}$ and $\mathbf{M_\mathrm{comp}'}$, yielding the final compositional result (PartField + RANSAC). This baseline isolates the contribution of our robust registration and collision-aware optimization design.
\end{enumerate}

\begin{figure*}[t]
    \centering
    \includegraphics[width=0.9\linewidth]{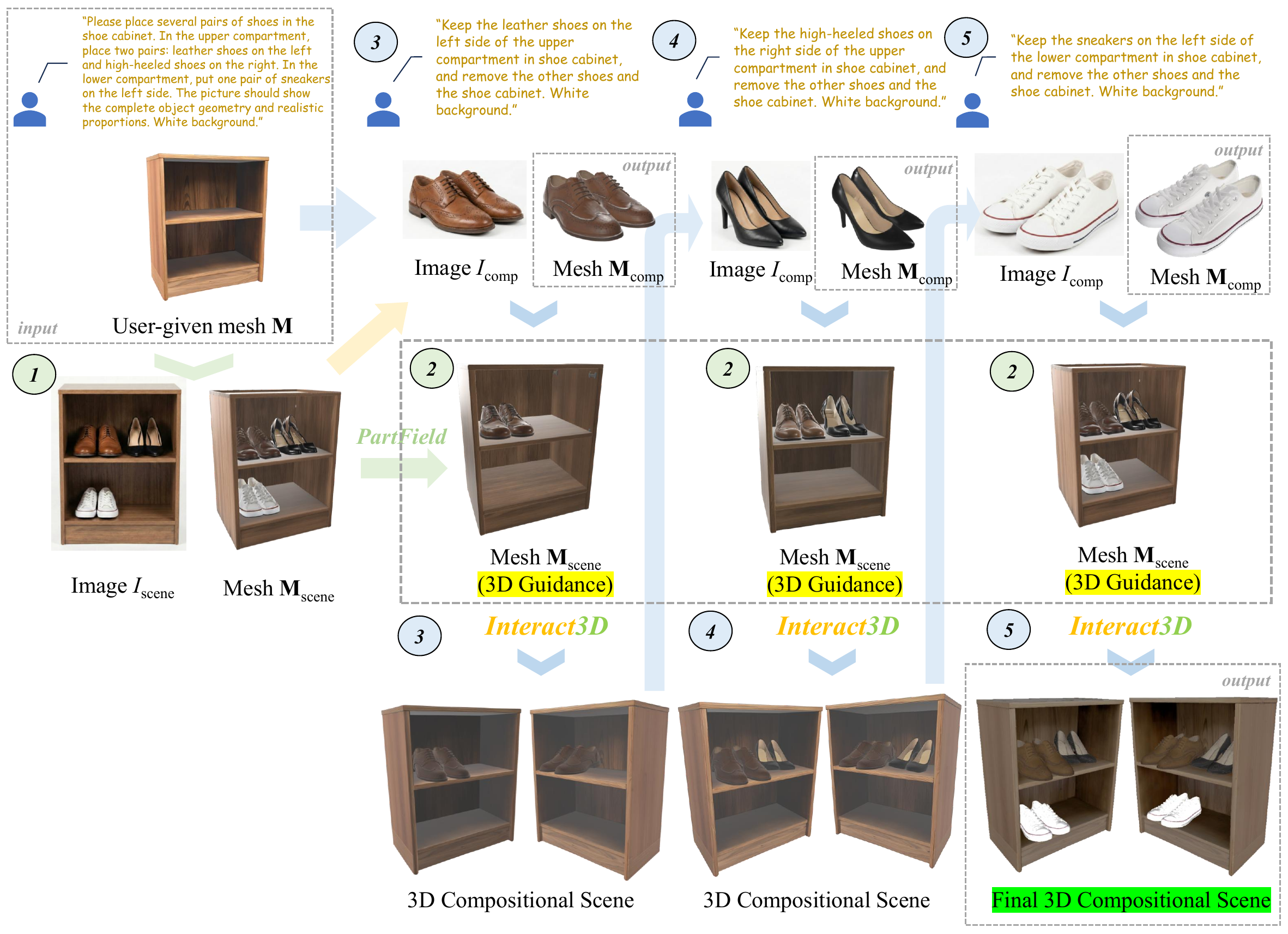}
    \caption{\textbf{More than two parts composition results.} Given a cabinet mesh, we render an image from it, add three pairs of shoes to the image using Nano Banana Pro and generate it with TRELLIS2 to obtain the object-object spatial relationships (OOR). Next, we use Nano Banana Pro again to extract each individual object in the shoe cabinet and generate them with TRELLIS2. Finally, using the OOR information, we sequentially add them to the cabinet mesh to form an interactive 3D scene.}
    \label{fig:ManyObjectsResults}
\end{figure*}

\subsection{Qualitative Evaluation}
\paragraph{\textbf{Two-part Composition.}} Figure~\ref{fig:QualitativeResults} presents qualitative comparisons of two-part composition results produced by different methods.
MIDI, trained on 3D-FRONT~\cite{fu20213d}, is inherently bottlenecked by the scarcity of 3D data, leading to poor generalization in generating individual 3D assets.
Similarly, both Jigsaw and 2BY2 rely heavily on their specific training data distributions, which limits the generalization capability of their position predictions. Consequently, their compositional outputs often exhibit severe misalignments or implausible spatial configurations.
Directly applying PartField to segment scene meshes generated by TRELLIS2 introduces visible geometric artifacts and structural holes, as the generated geometries are typically fused.
Moreover, the RANSAC-based alignment procedure struggles to achieve accurate geometric registration and may produce incorrect poses, such as the inverted pen inside the pen holder shown in rows 1 and 2.
Finally, independently registering the two objects fails to explicitly account for collisions, leading to severe geometric interpenetrations in the final compositions, as observed in rows 3, 4, 5, 6, 9, and 10.

\paragraph{\textbf{More-than-two-part Composition.}} Our framework naturally extends to compose an arbitrary number of objects, enabling the generation of rich interactive scenes. Similarly to our standard pipeline, after the user provides an initial 3D mesh, we first render it into an image and use Nano Banana Pro to edit the rendered image and introduce additional objects in the image space. We then use TRELLIS2 to generate a corresponding 3D scene. This provides spatial relationship priors that guide the placement of all components. Our multi-part composition is achieved through sequential two-object composition. As illustrated in Figure~\ref{fig:ManyObjectsResults}, the
user-provided mesh $\mathbf{M}$ is first treated as the anchor object $\mathbf{M}_{\mathrm{anchor}}$. A complementary object (e.g., leather shoes) is generated and treated as the remaining component $\mathbf{M}_{\mathrm{remain}}$, which is aligned to the anchor using our composition pipeline. The resulting composed scene is then treated as a new anchor, and another object (e.g., high-heeled shoes) is introduced as the remaining component. This process is repeated until all objects are incorporated into the final scene. More cases are provided in Appendix.

\subsection{Quantitative Evaluation}

We quantitatively evaluate composition results across 150 test cases (including cases in Figures~\ref{fig:QualitativeResults}, 7, and 10) based on two key aspects: the semantic fidelity, and the physical validity (see Table~\ref{tab:quantitative_results}). 

\begin{table*}[t] 
    \centering
    \caption{
        \textbf{Quantitative comparison of 3D compositional generation.} 
        We conduct a comprehensive evaluation using several unitless metrics, including semantic alignment measured by text- and image-based CLIP scores, human and VLM-based ratings, and geometric intersection rates computed over both surface and volume. The symbol ``/'' denotes metrics that are not applicable to a given baseline. The best results are highlighted in \textbf{bold}.
    }
    \label{tab:quantitative_results}
    \setlength{\tabcolsep}{10pt} 
    \resizebox{\linewidth}{!}{
    \begin{tabular}{lcccccc}
        \toprule
        & Jigsaw & 2BY2 & PartField & PartField+RANSAC & MIDI & \textbf{Ours}\\ 
        \midrule
        Text CLIP (Avg) $\uparrow$ & 0.3004 & 0.2798 & / & 0.3094 & 0.3079 & \textbf{0.3314} \\
        Image CLIP (Avg) $\uparrow$ & 0.7338 & 0.6982 & / & 0.7910 & 0.7840 & \textbf{0.8236} \\
        Human Rating (Avg) $\uparrow$ & 4.21 & 2.87 & / & 6.28 & 5.27 & \textbf{8.33} \\
        VLM Rating (Avg) $\uparrow$ & 5.89 & 4.08 & / & 6.59 & 5.84 & \textbf{8.57} \\
        $\mathbf{R_\mathrm{surface}}$ ($\times 10^{-3}$)(Avg) $\downarrow$ & 2.7792 & 1.5827 & / & 2.2287 & 22.185 & \textbf{0.6939} \\
        $\mathbf{R_\mathrm{volume}}$ ($\times 10^{-3}$)(Avg) $\downarrow$ & 16.837 & 9.1734 & / & 7.9280 & 116.70 & \textbf{3.8762} \\
        \bottomrule
    \end{tabular}}
\end{table*}

\paragraph{\textbf{Compositional Semantic Fidelity.}} Since PartField does not support compositional generation, our quantitative comparison focuses on the remaining four baselines. We use CLIP~\cite{radford2021learning} to measure the semantic alignment between 2D renderings of the final compositional 3D scene, captured from predefined viewpoints, and both the user-provided text prompt and the 2D image generated by Nano Banana Pro.
Higher CLIP scores denote stronger semantic consistency. 
To enable a more robust evaluation of OOR, we further include both human and VLM-based assessments using Qwen3, with ratings collected on a 1-to-10 scale.

\paragraph{\textbf{Geometric Intersection Rate.}} To evaluate collision avoidance, we measure interpenetration 
from both surface and volumetric perspectives, where lower values indicate better physical compatibility.

\emph{Surface Intersection Ratio.} We compute the total surface area involved in triangle–triangle 
intersections between meshes $A$ and $B$. Let 
$\mathcal{A}_{\mathrm{int}}$ denote the area of intersected surface 
regions, and $\mathcal{A}_A$, $\mathcal{A}_B$ the total surface areas 
of $A$ and $B$. The surface intersection rate is defined as:
\begin{equation}
\mathbf{R}_{\mathrm{surface}}
=
\frac{\mathcal{A}_{\mathrm{int}}}
     {\mathcal{A}_A + \mathcal{A}_B}.
\end{equation}

\emph{Volume Intersection Ratio.}
Since the compositional scene consists of solid objects represented by 
surface meshes, we estimate volumetric interpenetration via Monte 
Carlo sampling. Specifically, we uniformly sample $N(=10^6)$ points within 
a bounding box enclosing both meshes. 
The volumetric intersection ratio is defined as the fraction of 
sampled points that lie inside both objects among those that lie 
inside at least one object:
\begin{equation}
\mathbf{R}_{\mathrm{volume}}
=
\frac{\# \text{ points inside both } A \text{ and } B}
     {\# \text{ points inside } A \text{ or } B}.
\end{equation}

%% file: Section/6.Conclusions.tex
\section{Conclusion}
\label{sec:Conclusion}

We present \textbf{Interact3D}, a training-free generate-then-compose 
framework for interactive 3D scene synthesis. By leveraging spatial 
guidance from generative models, our method reformulates compositional 
generation as a collision-aware geometric alignment problem. Combining 
global-to-local registration, SDF refinement, and VLM-driven correction, our pipeline produces geometrically compatible and physically plausible scenes. We also release a curated compositional 3D dataset to facilitate future research in structured scene generation and robotic simulation.

\paragraph{\textbf{Future Work.}}
Although Interact3D generalizes well to daily objects, it struggles with fine-grained components (e.g., screws or tightly coupled joints). In these cases, severe occlusions in the 2D guidance images cause 3D geometric ambiguities, highlighting the inherent limitations of relying on 2D spatial priors for complex components. Future work will explore native 3D compositional generation to reason directly in 3D space, bypassing 2D dependencies and robustly handling intricate structures.

%% file: Section/Appendix.tex
\newpage
\section{Experiments}
% \paragraph{\textbf{Dataset Description.}}

\begin{figure*}[t]
    \centering
    \includegraphics[width=\linewidth]{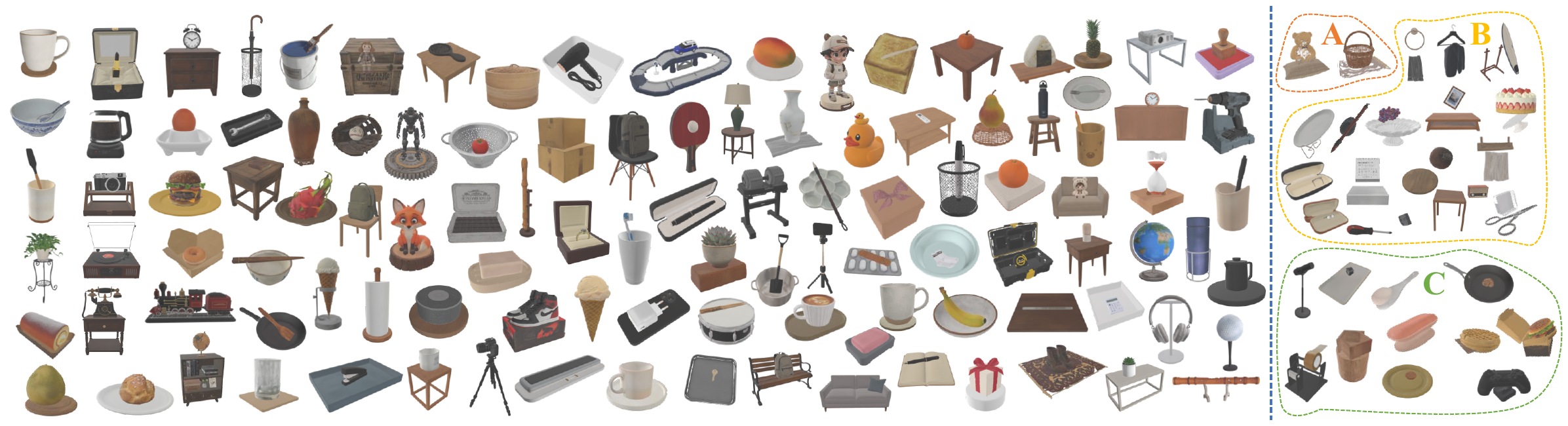}
    \caption{\textbf{Interact3D results on 140 additional cases} (left: success cases; right: failure cases). Only a small fraction of them exhibit failures: 2 cases (\textcolor{orange}{A}) suffer from mesh holes rooted in the TRELLIS2 reconstruction, 11 cases (\textcolor{ForestGreen}{C}) arise from symmetric ambiguity, and 16 cases (\textcolor{Goldenrod}{B}) result from collision-induced divergence.}
    \label{fig:140Results}
\end{figure*}

\subsection{Dataset Description}
Our dataset comprises approximately 8,300 compositional 3D scenes (sampled from about 10,000 candidates) spanning 9 categories: stationery, kitchenware, food, household items, electronics, apparel, decor, toys, and outdoor tools.
Specifically, about 7,700 scenes consist of two interactive objects, whereas the remaining ~600 scenes are composed of more than two objects (typically 3 to 5).
Each scene includes the individual 3D assets along with their corresponding compositional poses $\mathbf{\theta}$.
Notably, the user-provided reference meshes are flexible: they can be either high-fidelity assets obtained through manual 3D modeling or outputs synthesized by arbitrary 3D generative models (e.g., Hunyuan, TRELLIS2, Rodin).

\subsection{Ablation Study}

\begin{table*}[b] 
    \centering
    \caption{\textbf{Ablation study} of the proposed registration pipeline on 150 test cases. We compare RANSAC, ICP, their global-to-local combination, GeoTransformer (GeoT.), the SDF-based term, and the full agentic refinement loop. Best results are highlighted in \textbf{bold}.}
    \label{tab:AblationStudy}
    \setlength{\tabcolsep}{10pt} 
    \resizebox{\linewidth}{!}{
    \begin{tabular}{lccccccc}
        \toprule
        & RANSAC & ICP & RANSAC & GeoT. & GeoT.+ICP(SDF) & \textbf{Full}\\ 
        & only & only & +ICP & only  & w/o Agent & \\ 
        \midrule
        Text CLIP (Avg) $\uparrow$ & 0.3094 & 0.2985 & 0.3023 & 0.3187 & 0.3277 & \textbf{0.3314} \\
        Image CLIP (Avg) $\uparrow$ & 0.7910 & 0.7729 & 0.7958 & 0.8094 & 0.8101 & \textbf{0.8236} \\
        Human Rating (Avg) $\uparrow$ & 6.28 & 5.04 & 6.40 & 7.03  & 7.94 & \textbf{8.33} \\
        VLM Rating (Avg) $\uparrow$ & 6.59 & 5.88 & 6.31 & 7.21 & 8.18 & \textbf{8.57} \\
        $\mathbf{R_\mathrm{surface}}$ ($\times 10^{-3}$)(Avg) $\downarrow$ & 2.2287 & 1.2914 & 1.0481 & 0.8854 & 0.7458 & \textbf{0.6939} \\
        $\mathbf{R_\mathrm{volume}}$ ($\times 10^{-3}$)(Avg) $\downarrow$ & 7.9280 & 9.7102 & 6.7536 & 6.0039 & 4.9380 & \textbf{3.8762} \\
        \bottomrule
    \end{tabular}}
\end{table*}

We conduct ablation studies on our registration pipeline using 150 test cases (see Table~\ref{tab:AblationStudy}), primarily to validate the following three aspects. (i)~\textit{Global-to-local registration and GeoTransformer (GeoT.) robustness.} Pure ICP underperforms pure RANSAC, as it is prone to local optima. RANSAC+ICP, an alternative global-to-local scheme, yields improvements but still falls short of the GeoT.-based variants. (ii)~\textit{SDF term.} Removing it deteriorates penetration metrics, confirming its role in collision avoidance. (iii)~\textit{Agentic loop.} Removing it leads to consistent drops across all metrics, validating the effectiveness of the semantic-level corrections. (Here, ``RANSAC only'' corresponds to our baseline ``PartField+RANSAC''.)

\begin{figure*}[t]
    \centering
    \includegraphics[width=\linewidth]{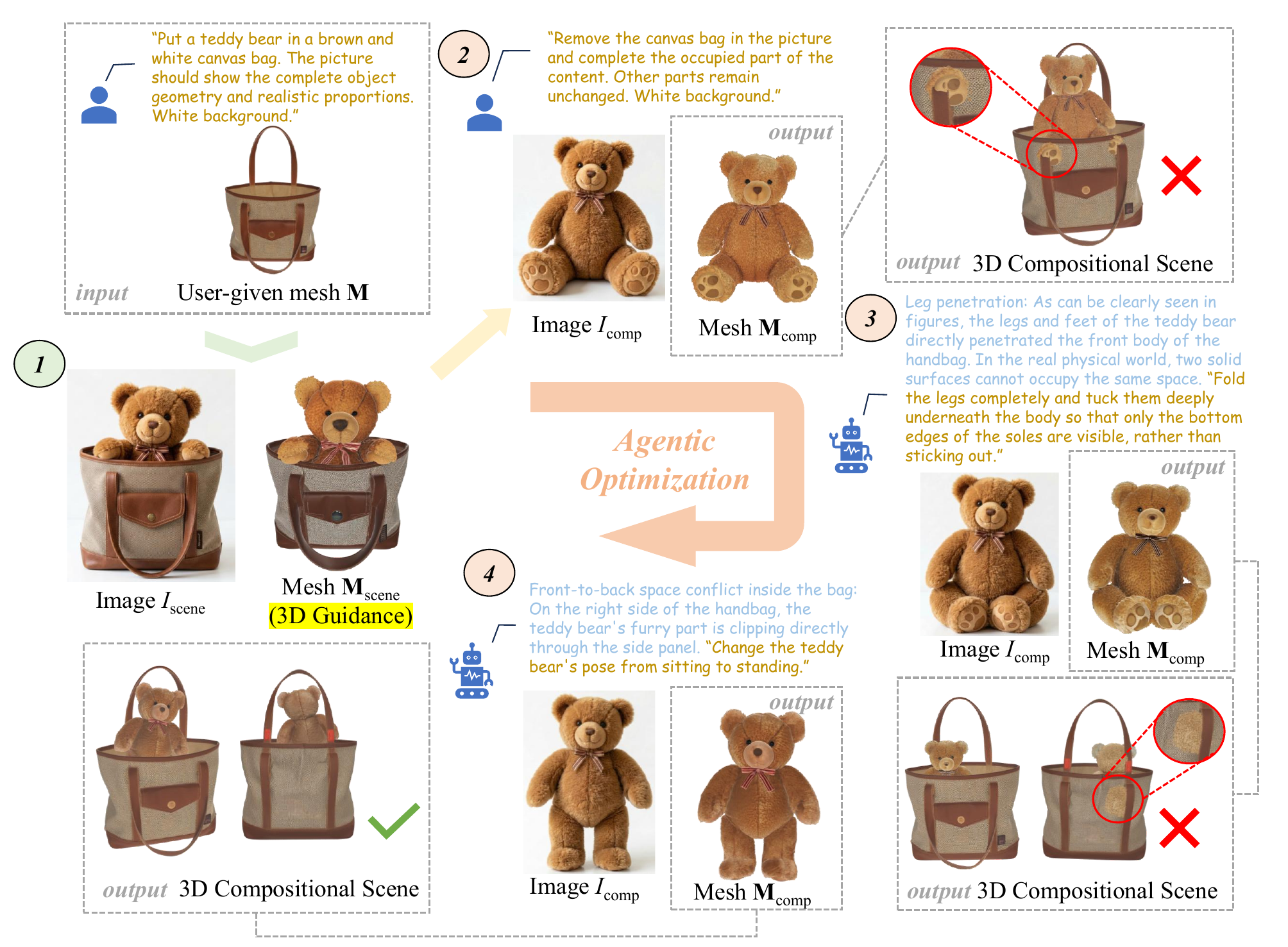}
    \caption{\textbf{Visualizations of Agentic Optimization.} Progressive geometric refinement of a teddy bear to achieve a collision-free fit inside a canvas bag.}
    \label{fig:AgenticOptimization_Sup}
\end{figure*}

\subsection{More Results}
\paragraph{\textbf{Effectiveness of Agentic Optimization.}} Severe 2D occlusions inevitably lead to collisions in the final 3D scene. As demonstrated in the additional case (Figure~\ref{fig:AgenticOptimization_Sup}), our agentic optimization effectively resolves these conflicts by generating an updated complementary mesh, ensuring a physically plausible final composition.

\begin{figure*}[t]
    \centering
    \includegraphics[width=\linewidth]{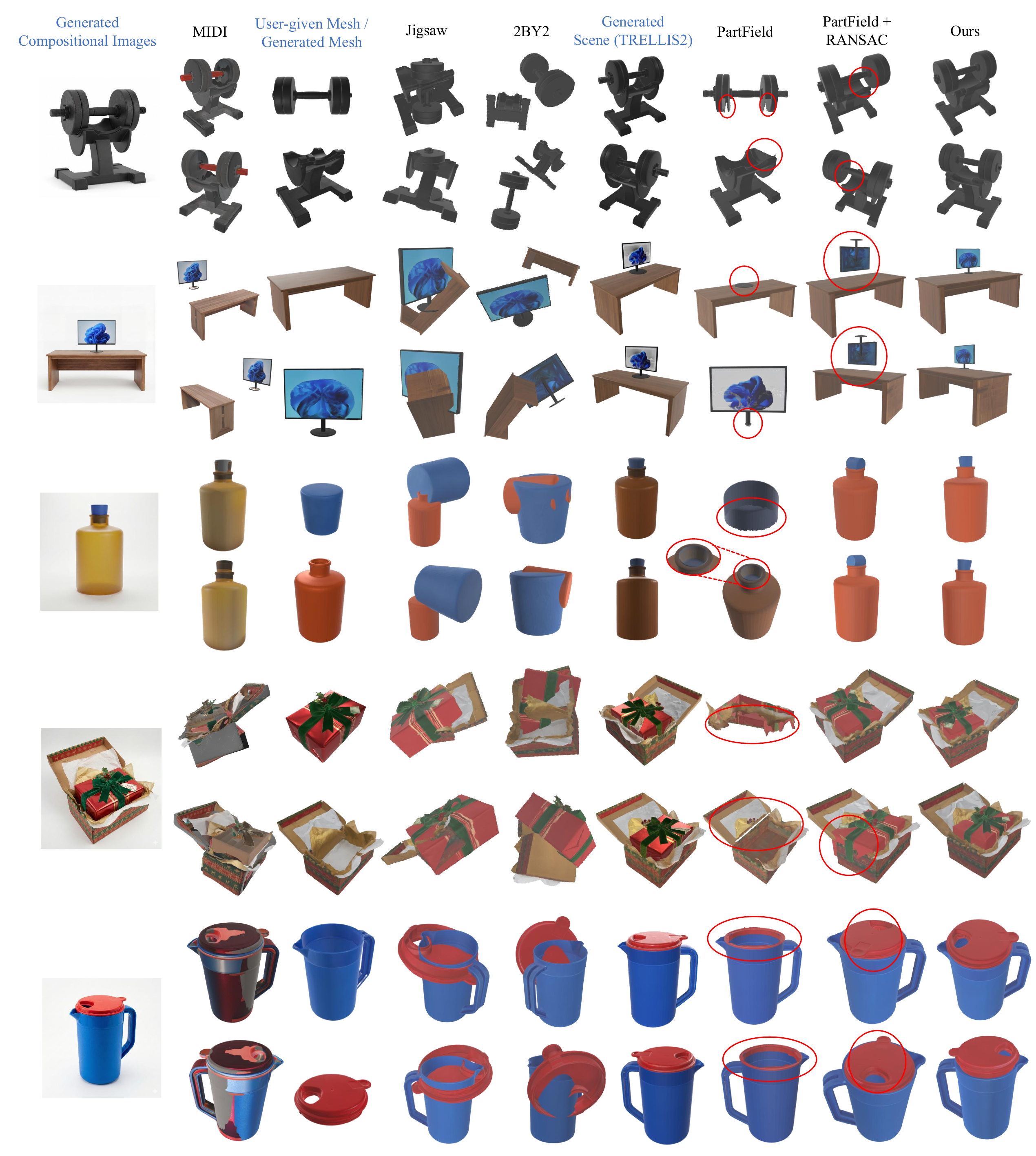}
    \caption{\textbf{Additional qualitative comparison.} 
    Interact3D outperforms MIDI, Jigsaw, and 2BY2 in terms of both individual asset quality and spatial position prediction, while also surpassing PartField in asset geometric fidelity.
    Furthermore, unlike the ``PartField + RANSAC'' baseline, our method inherently guarantees physical plausibility and precise spatial alignment.}
    \label{fig:Qualitative2}
\end{figure*}

\paragraph{\textbf{Two-part Composition Qualitative Evaluation.}} Figure~\ref{fig:140Results} presents 140 additional results on Interact3D. Figure~\ref{fig:Qualitative2} presents a qualitative comparison of the five cases across all baselines. Compared to MIDI, Jigsaw, and 2BY2, Interact3D demonstrates superior generalization in terms of both individual asset quality and spatial position prediction. Furthermore, our generated 3D assets exhibit significantly higher geometric fidelity than those produced by direct segmentation in PartField. When evaluated against ``PartField+RANSAC'', Interact3D showcases more robust physical plausibility and spatial alignment capabilities. Notably, in the 3rd and 4th rows, the baselines struggle with severe artifacts, resulting in unrealistic upside-down inversions of the computer.

\begin{figure*}[t]
    \centering
    \includegraphics[width=0.8\linewidth]{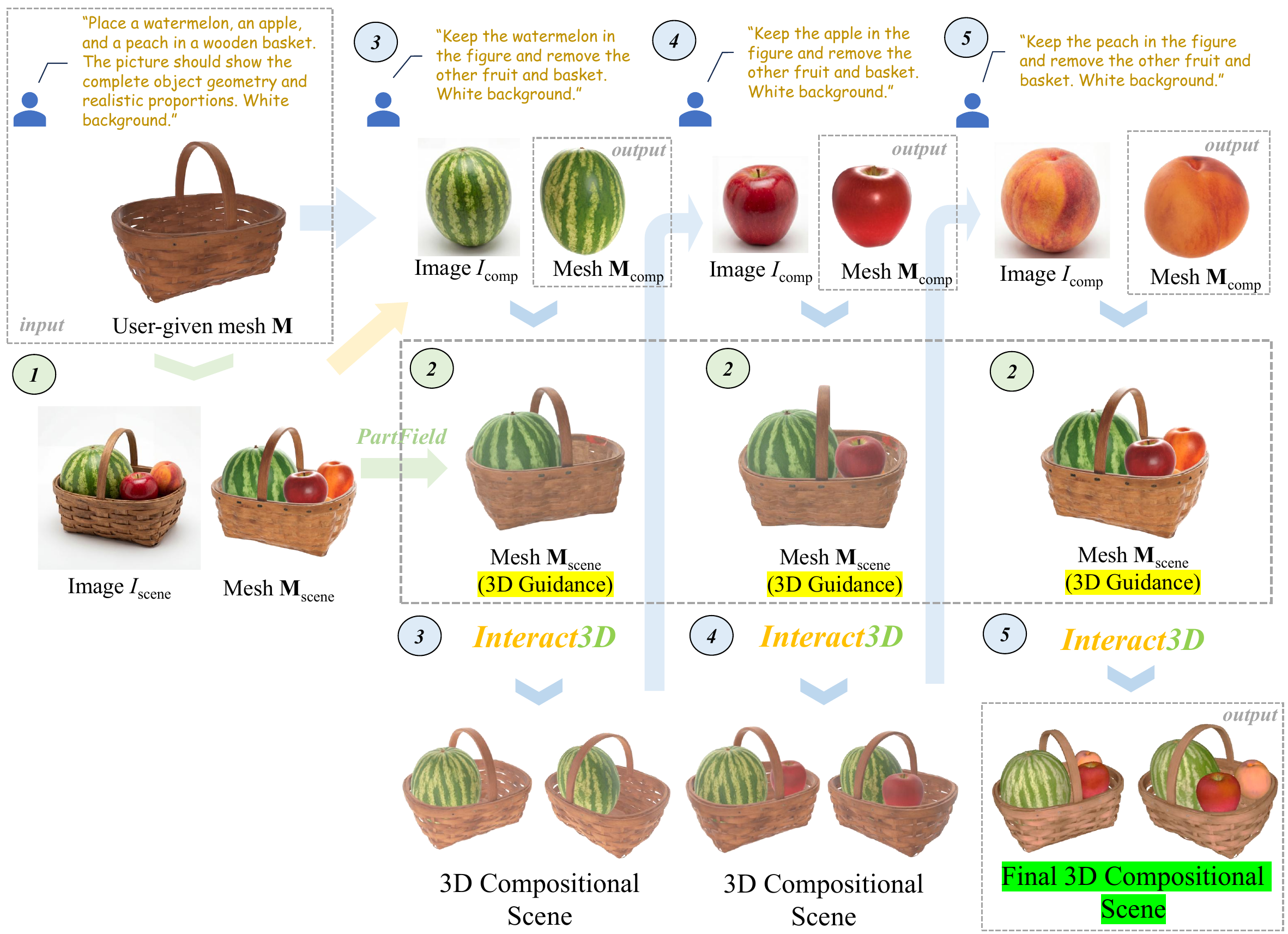}
    \caption{\textbf{Multi-part composition.} A user-provided basket is rendered and populated with fruits via Nano Banana Pro, then reconstructed by TRELLIS2 to extract object-object spatial relationships (OOR). Individual fruits are similarly synthesized and sequentially registered into the basket using the OOR guidance, forming a highly coherent, interactive 3D scene.}
    \label{fig:sup_manyobjects1}
\end{figure*}

\begin{figure*}[t]
    \centering
    \includegraphics[width=0.8\linewidth]{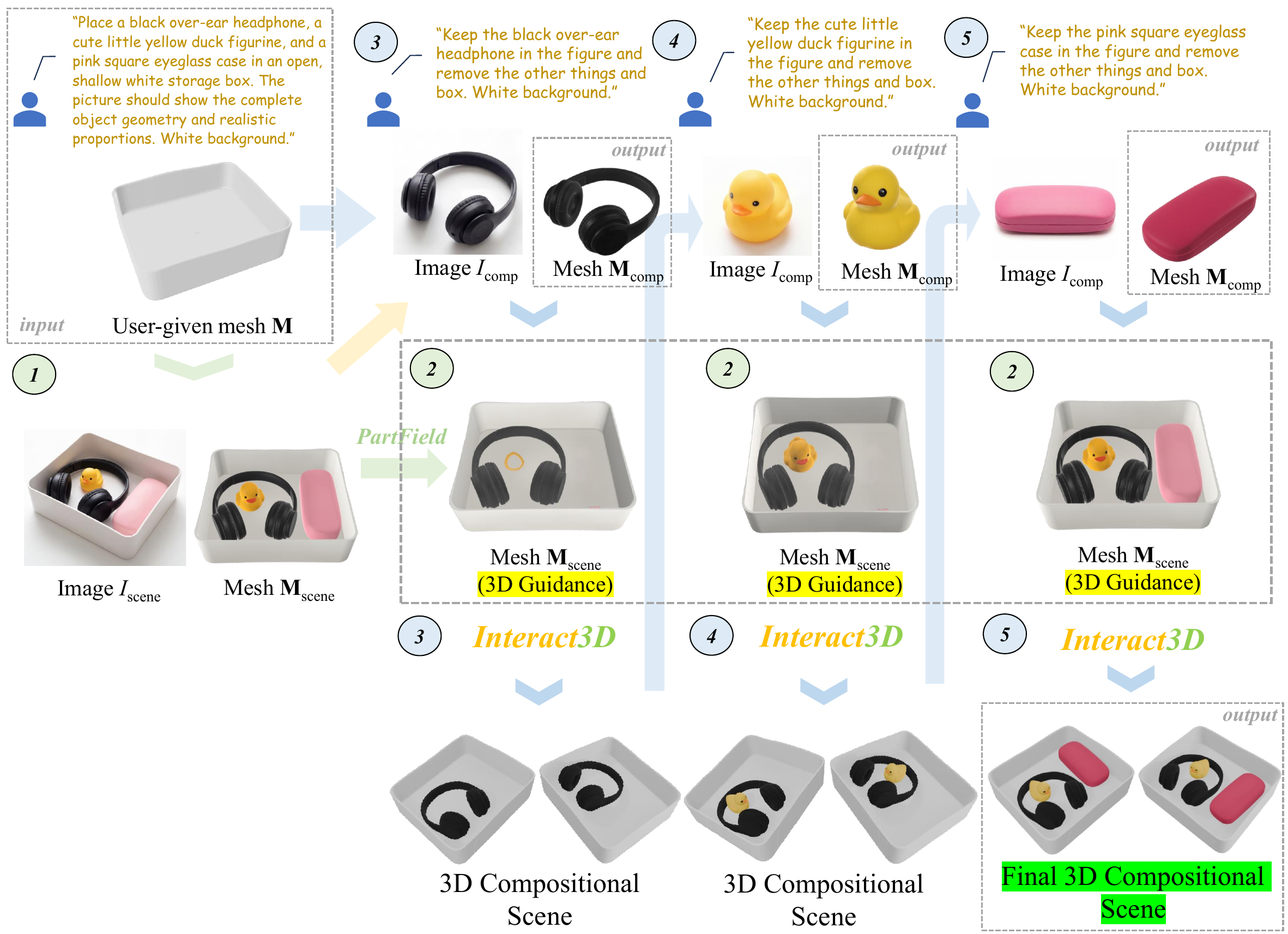}
    \caption{\textbf{Multi-part composition.} A rendered box is populated with a headphone, duck, and eyeglass case via Nano Banana Pro, then reconstructed by TRELLIS2 to extract object-object relationships (OOR). Individual items are then independently synthesized and sequentially registered into the box using this OOR guidance.}
    \label{fig:sup_manyobjects2}
\end{figure*}

\paragraph{\textbf{More-than-two-part Composition.}} To further validate the scalability and robustness of Interact3D, we evaluate its performance on more complex multi-part composition tasks (more than two parts), as visualized in Figures~\ref{fig:sup_manyobjects1} and~\ref{fig:sup_manyobjects2}. Although compositional scenes with more parts inherently introduce compounded spatial conflicts and severe occlusions, our framework consistently yields highly coherent compositions. Even in these highly constrained scenarios, Interact3D successfully maintains precise spatial alignment and strict physical plausibility across all constituent assets.

\begin{figure*}[t]
    \centering
    \includegraphics[width=0.9\linewidth]{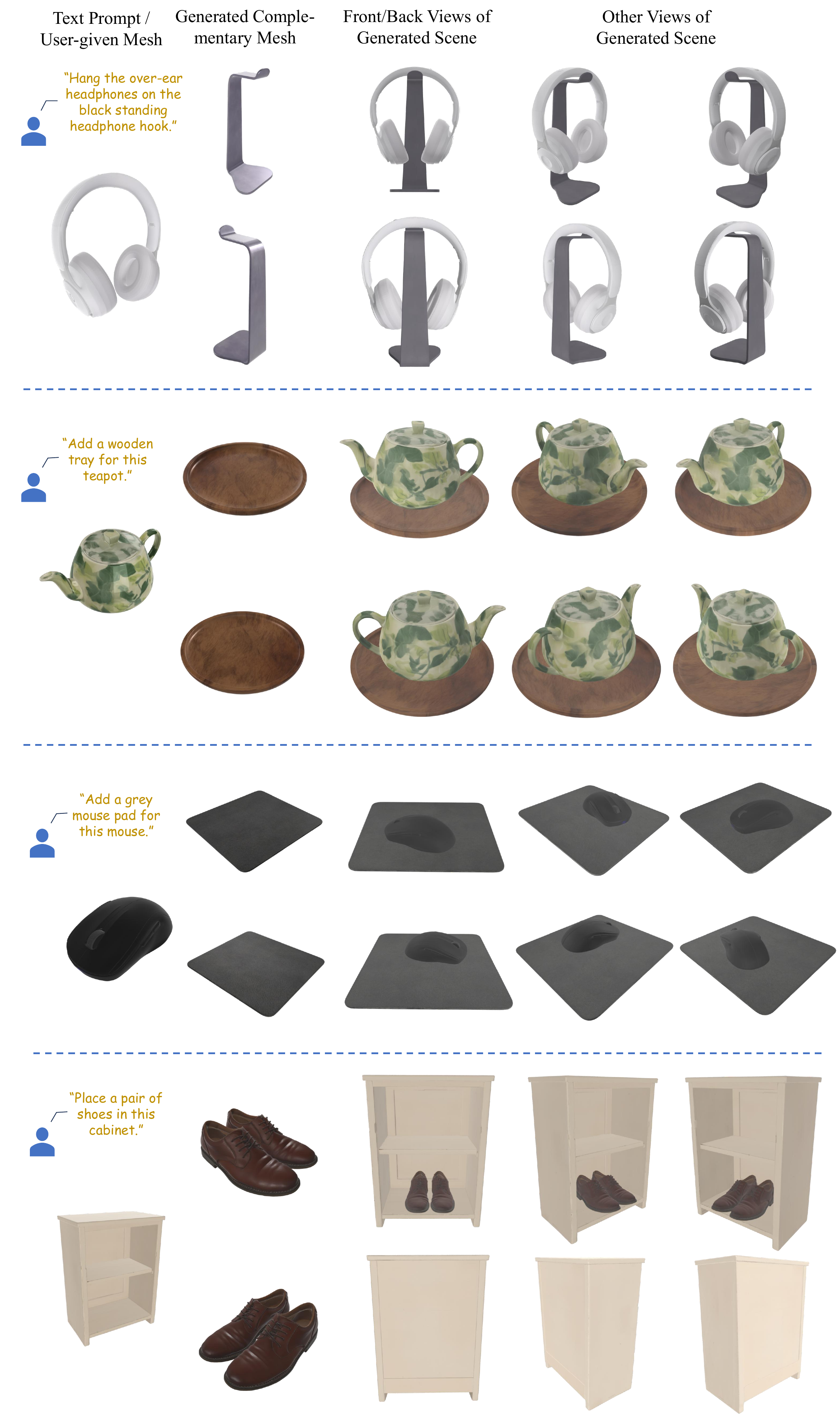}
    \caption{\textbf{Multi-view rendering results.} Given a user-provided mesh and a textual prompt, Interact3D synthesizes a complementary mesh and composes them into a coherent, interactive 3D scene. We visualize the compositional scenes rendered from six distinct viewpoints.}
    \label{fig:Sup_Multiview1}
\end{figure*}

\begin{figure*}[t]
    \centering
    \includegraphics[width=0.9\linewidth]{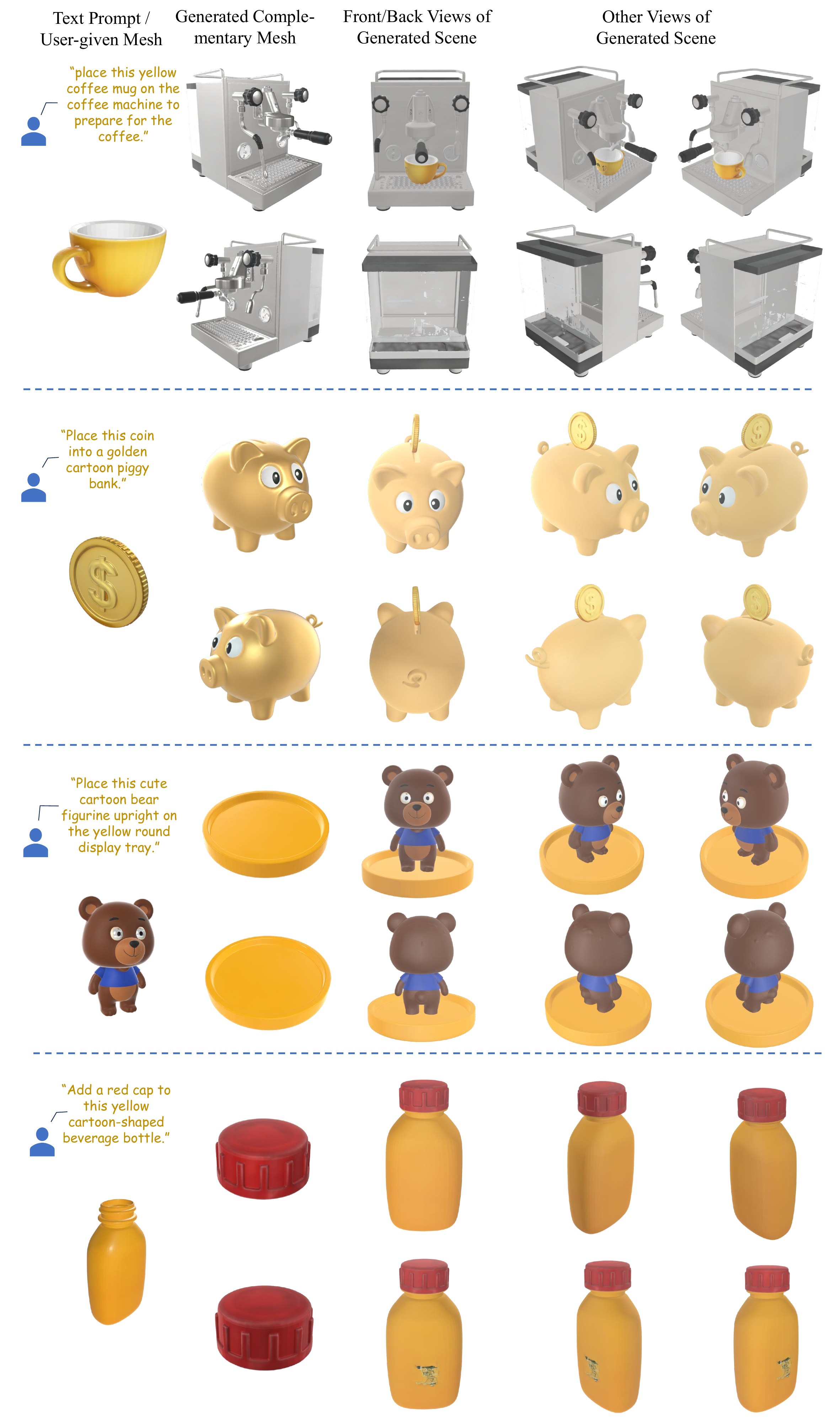}
    \caption{\textbf{Multi-view rendering results.} We visualize the compositional scenes rendered from six distinct viewpoints.}
    \label{fig:Sup_Multiview2}
\end{figure*}

\begin{figure*}[t]
    \centering
    \includegraphics[width=0.9\linewidth]{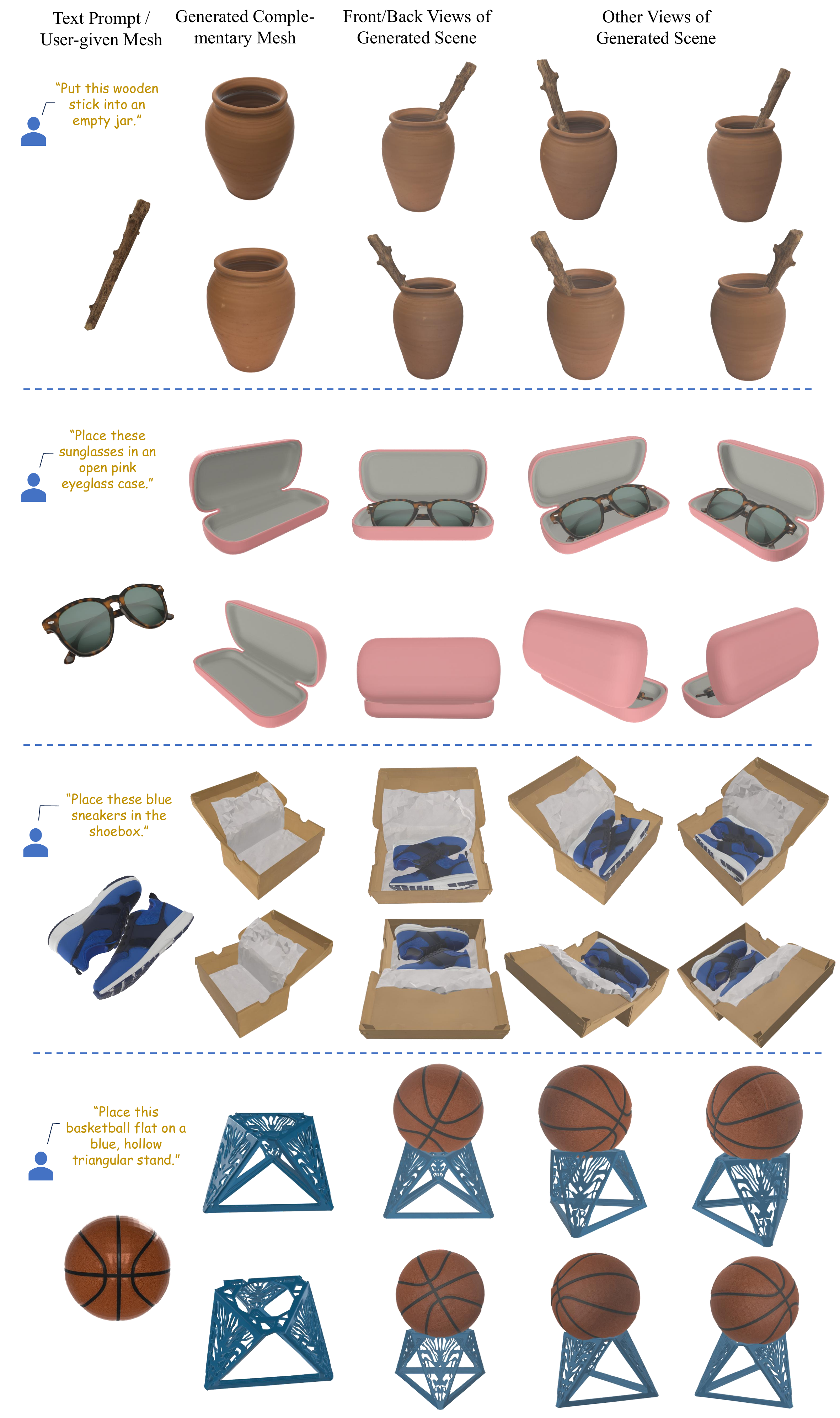}
    \caption{\textbf{Multi-view rendering results.} We visualize the compositional scenes rendered from six distinct viewpoints.}
    \label{fig:Sup_Multiview3}
\end{figure*}

\paragraph{\textbf{Multi-view Rendering Results.}} Since two-viewpoint evaluations may obscure geometric flaws, Figures~\ref{fig:Sup_Multiview1},~\ref{fig:Sup_Multiview2}, and~\ref{fig:Sup_Multiview3} showcase our compositional scenes from multiple viewing angles. This exhaustive multi-view display validates Interact3D's capability. It maintains the high quality of the constituent 3D assets and also enforces collision-aware constraints, resulting in physically plausible composition from any viewing angle.

\begin{figure*}[t]
    \centering
    \includegraphics[width=\linewidth]{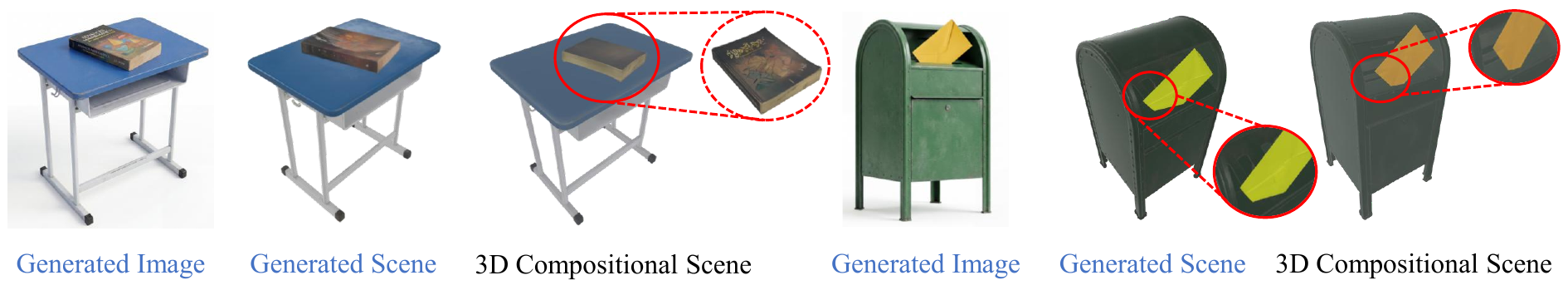}
    \caption{\textbf{Failure cases.} \textbf{(left)} Strong geometric symmetry in books can lead to upside-down orientations despite accurate spatial localization. \textbf{(right)} Heavy 2D occlusions cause severe mesh interpenetrations in the TRELLIS2-generated prior, propagating these physical collisions directly to the final compositional scene.}
    \label{fig:FailureCases}
\end{figure*}

\subsection{Failure Cases}
Because our registration in composition tasks relies on geometry-aware GeoTransformer without texture awareness, objects with strong geometric symmetry (such as cuboid books) can cause orientation ambiguity, occasionally leading to upside-down inversions (see Figure~\ref{fig:FailureCases} (left)). Additionally, since our composition pipeline is explicitly supervised by the spatial priors of $\mathbf{M_\mathrm{scene}}$, initial spatial misalignments generated in extremely complex scenarios can propagate downstream, ultimately compromising the final composition quality of Interact3D (see Figure~\ref{fig:FailureCases} (right)).

